TITLE: The Einstein Test: Towards a Practical Test of a Machine's Ability to Exhibit "Superintelligence"

AUTHORS: David Benrimoh[1,2*], Nace Mikus[3]  Ariel Rosenfeld[4]

Affiliations:

1. McGill University, Department of Psychiatry

2. Douglas Research Center, Montreal, Quebec, Canada

3. Interacting Minds Center, Aarhus University, Aarhus, Denmark

4. Bar-Ilan University, Israel

*Corresponding author contact: david.benrimoh@mcgill.ca

Abstract:

Creative and disruptive insights (CDIs), such as the development of the theory of relativity, have punctuated human history, marking pivotal shifts in our species' intellectual trajectory. Recent advancements in artificial intelligence (AI) have sparked debates over whether state-of-the-art models possess the capacity to generate CDIs. We argue that the ability to create CDIs should be regarded as a significant feature of machine superintelligence (SI).To this end, we propose a practical test to evaluate whether an approach to AI targeting SI can yield novel insights of this kind. We propose the "Einstein test": given the data available *prior to* the emergence of a known CDI, can an AI independently reproduce that insight (or one that is formally equivalent)? By achieving such a milestone, a machine can be considered to at least match humanity's past top intellectual achievements- and therefore to have the potential to surpass them.

Main Text:

*Introduction*

Current advances in artificial intelligence (AI) are proceeding at an unprecedented pace. The speed of this progress has prompted some to declare that not only will AIs match human intelligence in most or all domains in short order (which remains a controversial claim[1]), but that they will soon far exceed human intelligence in every domain, reaching a form of so-called superintelligence, or SI[2,3]. Clearly, the impact of SI on human society would be profound. SIs could theoretically revolutionize everything from medicine to governance; and they could potentially solve problems, such as the reconciliation of relativity with quantum theory, which have eluded humans thus far. The philosophical implications of this are fascinating[4] it is evident that many companies and researchers are optimistic about the benefits of SI and are actively pursuing its development[5], committing considerable financial, intellectual, and material resources in this effort[6]. At the same time, a true SI could create a disproportionate concentration of power in the hands of those who control it, and further exacerbate inequality and geopolitical instability[7,8]. Given how societies struggle to adapt to rapid technology changes, it is imperative to establish rigorous benchmarking systems to detect, evaluate, and regulate SI development.

Such benchmarks would serve a role similar to the ARC-AGI dataset in the quest for artificial general intelligence[9]. However, there are challenges in determining whether an AI is likely to develop SI. Specifically, we lack particular milestones that are expected to be achieved on the road to SI. Clearly, SI would surpass human intelligence, but identifying how and when this has occurred is far from straightforward. This complexity stems from two major issues: the lack of consensus on how to define and measure human intelligence, and the absence of a formal framework for determining when an AI has exceeded human capabilities across most domains. Some might argue that we will "know an SI when we see it", but this assumption is fraught with uncertainty. Without benchmarks or milestones, it may not be clear whether an AI has the potential to achieve SI until it already has. This ambiguity underscores the central challenge addressed in this article: defining a practical test that SI machines- or those likely to develop SI- may be expected to pass.

*Creative and Disruptive Insights*

We approach this challenge by considering "Creative and Disruptive Insights" (CDIs). These are, essentially, the kinds of insights that, prior to the development of machine intelligence, we would have considered some of the most remarkable exemplars of human intelligence. These are the paradigm shifts in our history by individuals or groups that have not only enabled new discoveries, but have completely changed our understanding of the world and led to significant developments [10]. To be classified as a CDI, the insight must be a step-change in the non-hyperbolic sense of the term, such that if one were to read the entire corpus of human knowledge before and after the CDI, one would be able to easily demonstrate that the concepts generated by the CDI were simply not present in an appreciable form prior to the CDI, and that after the CDI these concepts led to significant changes in human affairs.

One example that comes readily to mind is the development of relativity theory by Einstein; another example would be the development of calculus by Newton and Leibniz. We do not espouse the requirement that the CDI be the work of a prototypical 'solitary genius', nor do we require that the CDI have occurred rapidly or had a rapid impact- the development, by numerous humans collaborating over time, of a given field - for example, quantum mechanics - would reasonably be considered as a CDI. Importantly, we focus on CDIs that require creative and novel thinking. In other words, CDIs discovered by a matter of chance (e.g. the discovery of penicillin) or exhaustive search (e.g. the periodic table) are not of particular interest as far as SI is concerned, both because this would strain the definition of intelligence and may, in some cases, be prohibitive in terms of time of resources (see[11]). That being said, CDIs could potentially be related to the recognition of opportunities presented by chance events (but in those cases, the chance event would be the starting point, not the endpoint). One key element for our definition of the CDI is the capacity to set clear 'before' and 'after' points in time, for reasons which will be clear shortly.

Today, a machine may be able to speak by imitating humans via hyper-scaled large language models, without understanding what it says or how it 'reasons'[12]. As such, tests such as the Turing test[13] may no longer be effective in evaluating machine intelligence in general and SI in particular. However, if a machine were to, on its own, generate a CDI such as special relativity, then, we argue, we would be hard-pressed to dismiss its capacity for truly creative and transformative thought, without first calling into question what we believe is special about human intelligence. Therefore, in our conceptualization, the greatest CDIs of human history *can be considered as a labelled dataset on which machine intelligence can be tested* to determine if it has the capacity to develop SI by demonstrating that it can, at least, match clear instances of past top celebrated human intelligence that transcends 'normal' human intelligence.

Before continuing, we note that we choose to focus on CDIs from the fields of physics and mathematics because these tend to be provable and verifiable, such that if a machine intelligence were to arrive at the same idea but via different means, analogies, or formalisms, the fundamental point(s) being made should be formally equivalent to the original human-generated CDI. We endorse the idea that CDIs can be as easily present in other sciences, in the arts, and commerce; in these disciplines, however, we anticipate it would be more challenging (and perhaps less meaningful in the case of art) to demonstrate formal equivalence between an AI-generated candidate CDI and the historical equivalent, though this could easily be the subject of future discussion.

*The "Einstein Test of SI Candidacy"*

We propose the following test: given all human knowledge available prior to the development of an established CDI, can a machine develop a CDI at least equivalent to the one developed by humans?

Note that the machine has an inherent advantage in this formulation as no human ever could process all of the available knowledge at a given moment in human history. As such, the AI will, in practical terms, have access to more information than the CDI progenitors.

As a practical example, a dataset could be constructed of all information available prior to the publication of special relativity in 1905[14]. An AI 'contestant' would then be granted access to this data, and instructed to try and resolve the questions which prompted Einstein to pursue his work[14]. Success would be achieved if the AI generated a set of ideas that are formally equivalent to or encompass those contained within the theory of relativity. Failure would be an incomplete or incorrect solution, or the determination by the system that it cannot solve the challenge.

This example of what we can call, perhaps, the "Einstein Test of SI Candidacy", raises several important points regarding the precise construction of the test which will be necessary to avoid spurious results, while also giving the AI a realistic chance of succeeding. For example, many CDIs were arrived at over time and in the context of experimentation- the results of which would not be available in the training data. As such, it would only be reasonable that the AI should be provided the results of experiments it requests be conducted, so long as these would have been possible in the context of the time period (if not, the AI would need to make the necessary inventions to render the experiment possible). In addition, humans have access to a wealth of data from sense-organs and interactions with other humans which are not part of knowledge databases; as such, AI architectures with embodied components, which can take in real or simulated sensory data, or networks of AIs working in teams (or as adversaries), would be acceptable.

Another key point is the appropriate motivation of the system. As users of modern large language models (LLMs) know, the construction of prompts for an LLM is a key determinant of the quality and nature of the response. However, in the context of the Einstein Test, care will need to be taken that the prompt or equivalent motivation given to the AI does not include hints at the answer. On the other hand, some kind of motivation is necessary, otherwise the AI will not know what to focus on. As such, some forensic work will be required by science historians to provide the questions that motivated the progenitors of the CDI *prior to them knowing what it was they would discover or invent.* Indeed, it will be important to ensure that the AI is not even given an impression that there *is* a specific answer to the initial question(s), to preserve its ability to fail by concluding no answer is possible (we argue that it is critical to allow an AI to be able to decide not to continue, otherwise it would potentially never stop trying to answer any given question). Similarly, it will be important to ensure that no bias is built into the AI training protocol which would artificially privilege certain data which would, in turn, give a hint towards the solution planted by the human developer rather than noticed by the AI.

The test would proceed as follows. First, the developers of the candidate AI would define a specification and training procedure for their AI. The training procedure would need to be reviewed by the committee administering the test in order to ensure that no bias is present.

Second, a CDI would be revealed by the committee. Crucially, if we want a true SI, then the AI should be able to function without being fine-tuned by human developers for specific scenarios.

Third, the AI would be provided with a curated dataset containing any and all data available prior to the development of the CDI. This dataset could be augmented by simulated or real sensory data, or more advanced options, such as a simulated developmental period. The key issue to be monitored by the committee during their preparation of the dataset, or while reviewing simulated sensory data to be included in the training data, is that no knowledge from the period of time *after* the CDI is included in any of the training data.

Fourth, the AI would be trained on the data; as noted, the training protocol would need to be verified in order to avoid bias or the interjection of a "guiding hand" that would nudge the AI towards the solution. The data would remain available to the AI during the challenge so that it can re-train itself as needed based on the motivations provided.

Fifth, the AI would be given a set of challenges or questions- translated to whatever format its architecture requires- that were unsolved prior to the CDI, but which do not contain the answer within them. For example, in the Einstein example, to explain time dilation or the behaviour of light and electromagnetic waves, without the insight of relativity theory. As noted above, this would require the involvement of science historians to ensure that motivations would be in line with what was known prior to the CDI and what motivated the CDI's developers, and that the instructions or challenges are phrased in a manner that does not presuppose the form of the solution.

Sixth, a separate team from those that developed the AI should respond to queries from the AI. For example, if the AI requests data from an experiment, the team would provide this data if this would have been possible in the context of the time period of the CDI; else, they would respond that such an experiment is not possible without further innovations (which the AI would need to supply). While there is no specific reason that a CDI would have needed to occur in a given time period, this insistence on providing data from experiments that could only have occurred in the time period is intended to avoid any situations in which the provision of data from more advanced experiments would provide an inadvertent hint at the CDI solution. However, the AI would be free to design novel experimental techniques or apparatus and request results using these. If ever the AI develops a novel experiment that is cost-prohibitive compared to what was required to arrive at the CDI historically, then it would be reasonable to deny it the results- such an expensive experiment was not required for the original CDI, and the AI needs to demonstrate a capacity to adapt and manage resources in any case if it is to develop into an SI.

The AI would then be allowed to run until it believes it has accomplished the task, provides an incorrect answer that it asserts is correct, or it declares that it was unable to solve the challenge.

*Practical Considerations for the Einstein Test*

Realistic time or resource limits (in terms of some combination of time compute power, and resources for experiments) should be introduced in order to make the test practical and useful, though we acknowledge that determining what 'reasonable limits' are might be a controversial matter and may change over time given advances in computational power. Resource limits would be important to consider because if a machine requires significantly more resources to develop a CDI than its human progenitors, then it suggests further optimizations to the machine intelligence may be possible. Clearly, therefore, the allotted resources would need to scale with the specific CDI. In this context, the time it takes the human team to produce any experimental results during the sixth stage should be carefully considered if these results are not available from simulations or the historical record.

Beyond hard resource limits, having the machine decide when it cannot resolve a problem would be a potentially more elegant way to identify failure states. To be clear, we are not discussing a situation where a machine would "prove" as this would be problematic[15]; rather, we mean that the machine would be capable, as a human would, of having an intuition as to whether or not its effort was likely to be fruitful. In addition, this may teach us different things about the machine's reasoning strategy compared to a completely false result. While not required for the competition, it may also be beneficial to inspect the intermediate states of the AI in order to assess if the AI has 'come close' to the CDI; however, measuring 'closeness' in this context would pose a significant challenge as progress towards a CDI should *a priori* be expected to be linear. As an additional consideration, some of the CDI benchmarks could go beyond binary success definitions and provide continuous measures of the "closeness" to the solution (e.g. how close the machine gets to predicting a physical quantity, such as an orbit or particle mass, which would be predicted from the theoretical advancements inherent in the CDI).

The application of the Einstein Test and its constituent CDI cases would require a committee of experts in relevant fields (such as mathematics or physics), the history of science, and artificial intelligence. We do not lay out the technical specifications of any given CDI in this paper for this reason- we believe a multidisciplinary effort would be required for this, beyond the scope of the present viewpoint. We do recommend a call for CDI proposals from the scientific community which would need to encompass the following: 1) the description of a dataset available at a justified timepoint relevant to the CDI; 2) a set of inciting problems or challenges not resolved by knowledge available in this dataset; 3) a suggested timeframe and compute limitation and 4) a benchmark for when the CDI is achieved (including equivalence statements, or benchmarks).

We would recommend that this committee's work be divided into two phases. In the first phase, they would need to develop the CDIs (or evaluate CDIs proposed by the scientific community) and oversee a competition aimed at encouraging development teams to accomplish them. Once some of the CDIs have been achieved, we would recommend that the committee move into a prospective competition. The winning AI(s) would be trained on all *currently* available data, and given a series of motivations to solve pressing global problems. This phase would have two objectives. The first would be to determine if the candidate AIs can solve problems of the same type as they were tested on (i.e., mathematics or physics), where solutions can be verified. The second would be to see if the candidate AIs can generalize to other problems- such as political or biomedical challenges, where potentially more than one correct solution exists.

As of today, we believe that achieving even one CDI would be a remarkable feat for an AI, and achieving, say, 5 or 10 of them would be extraordinary and would likely only be possible for an AI that would provide great utility even if it does not develop 'true' SI. This would be an advantage given the cost involved in developing current testing sets for artificial general intelligence efforts.

It is important to note that passing the above Einstein Test would strongly indicate the possibility of machine SI- as it will have at least matched the greatest instances of human insight and creativity. However, we cannot rule that an AI failing the test will *never* develop SI. Specifically, human-level intelligence has existed for, by definition, the entirety of human history- including the long stretches of it where the total amount of data available would have been vastly inferior to what is available today. However, AI ability to present human or above human intelligence benefits greatly from high quality large datasets. In many (though not all[16]) cases, the availability of large datasets is even required to achieve high performance. As such, it may be the case that an AI which failed the test would still be capable of producing CDIs given today's available data and would eventually develop SI. Such an examination could be performed in a prospective manner as suggested before, but would still present a less certain scenario as future CDIs are, by definition, unknown. Nevertheless, one may still argue that it is the architecture or method of construction of an AI which should hold the seeds of SI- in the same way that the human brain can clearly and reliably develop human-level intelligence and proficiency in specific fields without requiring the scale of human-produced data consumed by the most modern AI systems. Human brains do have access to vast datasets- particularly the large amount of naturalistic data from sense organs. That being said, our sense organs have been unchanged throughout history, so an AI possessed of a simulated sensorium, if it is to develop SI, could reasonably be expected to at least match human intelligence with amounts of human-produced data on par with what a human would consume. In addition, the sheer variability of data consumed by humans who all develop human level intelligence, and the relatively small amount of human-produced data required for humans to develop proficiency in a given field compared to an AI system, suggests that something in the blueprint of the human brain is relevant to its ability to generate human-level intelligence. If this is the case, then one may argue that it is the construction of the human brain and the *manner* in which it processes information and learns from its environment, rather than access to an arbitrary amount of data, which results in human-level intelligence. For all these reasons we believe that it is reasonable to expect that at least a subset of AI systems that will develop SI will also pass the outlined Einstein test, even if the available data for training for a given CDI challenge is inferior in volume and/or quality than what may be available to train AIs today.

*Conclusion*

The development of machines that can reproduce established and celebrated CDIs, as tested by the proposed Einstein Test, would, by definition, mean the creation of machines that can at the least match the creativity and impact of some of the greatest human minds in history. We argue that the development of AIs which can replicate these pinnacles of human achievement will present a strong signal for SI and teach us much about our own minds in the process. In particular, such AIs would likely have much to offer our species in terms of producing future CDIs. Our species, certainly, would have many interesting problems to offer them.

Acknowledgements:
We thank Prof. Karl Friston for his valuable comments on these ideas.

Disclosures: D.B. is a founder and shareholder of Aifred Health, which creates AI powered clinical decision support software for mental health clinicians; Aifred was not involved in and did not support this research.

Funding: D.B. is supported by a Fonds de Recherche du Québec Junior 1 clinician-researcher grant and holds a Sidney R. Baer Young Investigator Award from the Brain & Behavior Research Foundation.